\documentclass[letterpaper]{article} 
\usepackage[preprint]{aaai2027}  

\usepackage{amsmath}
\usepackage{amssymb}
\usepackage[hyphens]{url}  
\usepackage{graphicx} 
\usepackage{natbib}  
\usepackage{caption} 
\usepackage{algorithm}
\usepackage{algorithmic}

\usepackage{newfloat}
\usepackage{listings}
\DeclareCaptionStyle{ruled}{labelfont=normalfont,labelsep=colon,strut=off} 
\floatstyle{ruled}
\newfloat{listing}{tb}{lst}{}
\floatname{listing}{Listing}

\usepackage{booktabs}

\title{MeshPriorDiT: Hierarchical Modeling for Action-Conditioned Cloth Dynamics}
\author {
    Zihang Wang\textsuperscript{\rm 1,\rm 2},
    Jianming Hu\textsuperscript{\rm 1},
    Shang Su\textsuperscript{\rm 1,\rm 2},
    Hao Huang\textsuperscript{\rm 1},
    Mengkai Shi\textsuperscript{\rm 2},
    Jun Gao\textsuperscript{\rm 3}\corresponding,
    Shuo Feng\textsuperscript{\rm 1}\corresponding
}
\affiliations {
    \textsuperscript{\rm 1}Tsinghua University\\
    \textsuperscript{\rm 2}Dense-AI\\
    \textsuperscript{\rm 3}University of Michigan
}

\begin{document}

\maketitle

\begin{abstract}
Action-conditioned cloth dynamics prediction requires both locally plausible deformation and long-range coordination. Existing approaches largely follow two paradigms. Mesh-based GNNs capture local physical responses through material connectivity. However, their finite message-passing range limits coordination between topologically distant regions, while autoregressive rollouts tend to accumulate prediction errors. Transformer-based dynamics models capture long-range interactions through global attention, but often operate without explicit material connectivity and must learn local topological responses directly from data. We propose MeshPriorDiT, a hierarchical dynamics model that decomposes future cloth motion into a structured mesh prior and a generative residual. An action-conditioned mesh GNN first predicts multi-step vertex displacements, yielding a reference trajectory that respects material topology and grasp constraints. Conditioned on historical states, planned actions, and the mesh prior, a Residual DiT then uses conditional flow matching to jointly generate the residual motion not captured by the prior. The generated residual is further rescaled and decoded using material adjacency to coordinate corrections across neighboring vertices. We evaluate MeshPriorDiT on 15-step autoregressive rollout across three cloth manipulation tasks. Averaged over the three tasks, MeshPriorDiT reduces average Global MSE by \(43.42\%\) relative to the GNN-Only baseline and by \(75.03\%\) relative to the DiT-DDPM baseline, while maintaining a favorable Edge-strain MSE comparable to that of GNN-Only.
\end{abstract}

\section{Introduction}
Cloth manipulation is fundamental to robotic handling of deformable objects, with applications in folding and model-based action planning \citep{li2019particledynamics,lin2021softgym}. Cloth motion is governed by interconnected vertices and affected by material connectivity, grasp boundaries, folding, and self-contact \citep{tian2025diffusion,longhini2025clothsplatting,liao2024senc}. Dynamics models must therefore maintain both locally plausible deformations and coordinated motion across distant regions. Otherwise, errors accumulated over multi-step rollouts may alter fold locations and contact configurations, affecting subsequent action selection \citep{pfaff2021meshgraphnets,sanchez2020gns,tian2025diffusion,zhang2024adaptigraph}.

Existing learned approaches largely follow two paradigms. Mesh-based GNNs represent vertices and material connections as graphs and learn deformation propagation through local message passing \citep{pfaff2021meshgraphnets,longhini2025clothsplatting,zhang2024adaptigraph}. Although explicit topology provides a useful inductive bias for local physics, finite message-passing depth restricts communication between distant mesh regions \citep{fortunato2022multiscale,grigorev2023hood}. Moreover, single-step graph models recursively consume their own predictions, causing errors to propagate through long-horizon rollouts \citep{sanchez2020gns,wurth2025robin}. Multiscale and hierarchical graphs extend the propagation range but do not directly eliminate rollout error accumulation.

Transformer-based and generative models instead capture cross-region motion through global interactions \citep{tian2025diffusion,shao2022tie,zhang2026clothtransformer}. However, when material connectivity is not explicitly encoded, global attention does not guarantee consistent responses between neighboring material vertices. Recent methods combine local and global modeling through implicit interactions, hierarchical graph diffusion, MeshGraphNet--Transformer processors, or latent mesh tokens \citep{shao2022tie,wurth2025robin,iparraguirre2026meshgraphnettransformer,zhang2026clothtransformer}. Although some explicitly retain graph or mesh representations, local structure and global interactions are generally integrated within a shared representation or unified predictor. Their contributions are consequently entangled, making it difficult to identify which motion is explained by the mesh structure, which errors are corrected by global modeling, and whether such corrections preserve local consistency.

We propose \textbf{MeshPriorDiT}, a hierarchical dynamics model that explicitly decomposes future cloth trajectories into an action-conditioned mesh prior and a generative residual. A mesh branch propagates local motion along fixed material connections and predicts multi-step vertex displacements under grasp constraints. Rather than serving as the final prediction, its rollout provides a structured reference trajectory. A Residual DiT with a spatiotemporal Transformer backbone \citep{peebles2023dit} then conditions on historical states, planned actions, and the prior trajectory to generate the residual through conditional flow matching \citep{lipman2023flowmatching}. Magnitude scaling and topology-aware decoding coordinate residual corrections across neighboring vertices. The decoded residual is added to the mesh prior, while hard constraints ensure that controlled vertices follow the planned actions. Unlike general prior--residual generative formulations \citep{kutsuna2025residualprior}, our decomposition assigns explicit roles to the two branches: the mesh prior captures local physical propagation, while the Residual DiT corrects motion associated with long-range coordination, complex deformation, and rollout errors.

Our contributions are:
\begin{itemize}
    \item We propose MeshPriorDiT, which uses an explicitly correctable prior--residual interface to combine topology-aware local propagation with global residual modeling for action-conditioned cloth dynamics.

    \item We evaluate trajectory accuracy and structural quality across three manipulation tasks and multi-step autoregressive rollouts. Over 15 steps, MeshPriorDiT reduces average Global MSE by $43.42\%$ over GNN-Only and $75.03\%$ over DiT-DDPM, while maintaining Edge-strain MSE comparable to GNN-Only. Per-action and per-horizon analyses further characterize the complementary roles of the two branches.
\end{itemize}

\section{Related Work}
\subsection{Mesh-Based Learned Dynamics Models}
Graph-based simulators represent physical elements as nodes and learn relational dynamics through message passing \citep{battaglia2016interaction,sanchezgonzalez2018physics,sanchez2020gns}. MeshGraphNets specializes this formulation to mesh vertices and material edges, predicting state increments through an encode--process--decode architecture \citep{pfaff2021meshgraphnets}. Multiscale extensions shorten propagation paths using coarse graphs, hierarchical representations, or bi-stride pooling \citep{fortunato2022multiscale,grigorev2023hood,cao2023bsms}. These models provide explicit topological bias, but local receptive fields and autoregressive updates can limit long-range coordination and propagate errors across rollout steps.

Cloth-oriented models further incorporate observations, material variation, and contact. Cloth-Splatting combines action-conditioned graph prediction with an RGB-supervised update, while AdaptiGraph adapts graph dynamics to material properties \citep{longhini2025clothsplatting,zhang2024adaptigraph}. Physics-guided garment simulators learn deformation from unsupervised or self-supervised objectives \citep{bertiche2021pbns,santesteban2022snug,bertiche2022neuralcloth}; SENC and ContourCraft target self-collision and multi-garment intersections \citep{liao2024senc,grigorev2024contourcraft}. Although these methods improve perception, material adaptation, and collision handling, their dynamics modules remain complete-state predictors or state priors rather than explicitly correctable local-motion priors.

\subsection{Generative and Hybrid Dynamics Models}
Generative models provide global modeling beyond local graph propagation. Diffusion Transformers process noisy states, whereas Flow Matching and rectified flow learn continuous velocity fields along probability paths \citep{ho2020ddpm,peebles2023dit,lipman2023flowmatching,liu2023rectifiedflow}. UniClothDiff applies Transformer diffusion to cloth state estimation and action-conditioned prediction, while TIE models nonlocal particle interactions through implicit-edge attention \citep{tian2025diffusion,shao2022tie}. Such models capture distant interactions, but without explicit structural inputs they must infer material neighborhoods and control boundaries from data.

Hybrid approaches incorporate geometry into global sequence models. Mesh-reduced methods combine graph-based compression with temporal attention or flow generation \citep{han2022meshtemporal,sun2023sequentialflow}, and Manifold-Aware Transformers inject geodesic surface relations into attention \citep{li2024manifold}. ROBIN combines hierarchical graphs with rolling diffusion, whereas the recent MeshGraphNet-Transformer and ClothTransformer preprints introduce global or latent-space Transformers \citep{wurth2025robin,iparraguirre2026meshgraphnettransformer,zhang2026clothtransformer}. These components still jointly predict the complete state rather than assigning local physical propagation and residual detail to separate modules.

Iterative and residual methods provide another route to correction: PDE-Refiner improves long neural-solver rollouts, learned contact corrections augment reduced deformable dynamics, and the Residual Prior Diffusion preprint models deviations from a coarse generative prior \citep{lippe2023pderefiner,romero2021contact,kutsuna2025residualprior}. These works do not address correcting an action-conditioned cloth-mesh rollout under explicit grasp constraints. MeshPriorDiT instead uses the graph rollout as a structured local-physics prior and assigns coordinated residual motion to Residual DiT, yielding a hierarchical and diagnostically separable dynamics model.

\section{Method}

\begin{figure*}[t]
\centering
\includegraphics[width=\textwidth]{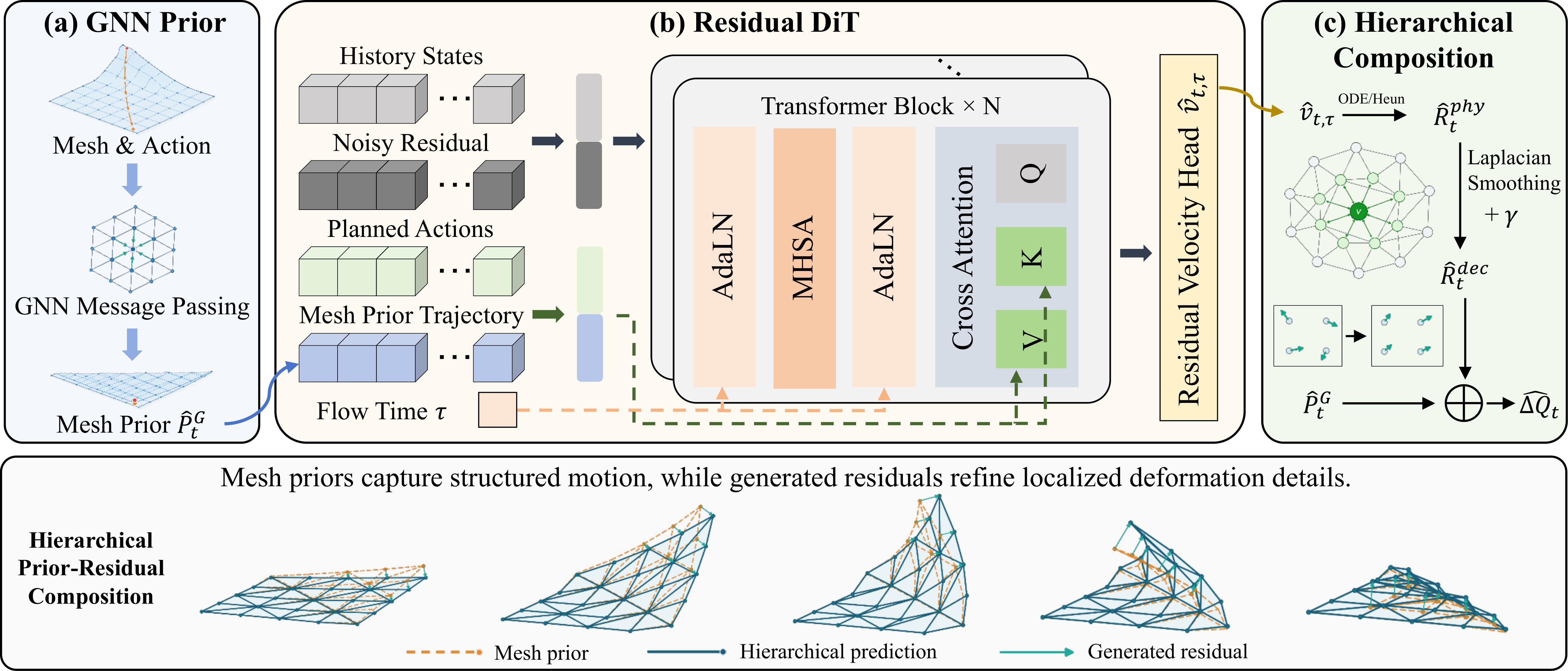}
\caption{\textbf{Overview of MeshPriorDiT.} (a) Given the cloth mesh and planned actions, the GNN propagates motion along material connections to generate an action-conditioned mesh prior capturing topology-constrained local dynamics. (b) Conditioned on the prior trajectory, Residual DiT predicts a velocity field for residual generation. (c) The generated residual is coordinated over material neighborhoods through graph-Laplacian smoothing, scaled by \(\gamma\), and composed with the mesh prior under grasp constraints. The bottom row illustrates the hierarchical decomposition. Orange dashed lines, blue solid lines, and green arrows denote the mesh prior, composed prediction, and generated residual, respectively.}
\label{fig:overview}
\end{figure*}

\subsubsection{Problem Setup.}
We study action-conditioned cloth dynamics prediction from full mesh states. The cloth is represented by a fixed template mesh
\(\mathcal G=(\mathcal V,\mathcal E,\mathbf U)\), where \(\mathcal V\) and \(\mathcal E\) denote the vertex set and material connections, respectively, \(\mathbf U\in\mathbb R^{N\times2}\) contains the two-dimensional material coordinates, and \(\mathbf q_t\in\mathbb R^{N\times3}\) denotes the vertex positions at time \(t\).

Given the most recent \(K\) states and a sequence of planned actions over the next \(H\) steps,

\[\mathbf Q_t^-=[\mathbf q_{t-K+1},\ldots,\mathbf q_t],\
\mathbf A_t=[\mathbf a_{t+1},\ldots,\mathbf a_{t+H}],\]

where \(\mathbf a_{t+h}\in\mathbb R^3\) denotes the end-effector displacement during the \(h\)-th prediction step. Each prediction window is also associated with a fixed grasp mask \(\mathbf M\in\{0,1\}^{N\times1}\), where \(M_i=1\) indicates that vertex \(i\) is directly controlled by the gripper. We use \(K=3\) and \(H=5\).

The model predicts future motion in the displacement space. The ground-truth target is defined as

\[\Delta\mathbf Q_t^*=[\Delta\mathbf q_{t+1}^*,\ldots,\Delta\mathbf q_{t+H}^*],\
\Delta\mathbf q_{t+h}^*=\mathbf q_{t+h}^*-\mathbf q_{t+h-1}^*,\]

where \(\mathbf q_t^*\equiv\mathbf q_t\). The corresponding prediction problem is written as

\[p_\theta\left(\Delta\mathbf Q_t\mid\mathbf Q_t^-,\mathbf A_t,\mathcal G,\mathbf M\right).\]

Future vertex positions are recovered by cumulatively applying the predicted displacements from the current state \(\mathbf q_t\).

\subsubsection{Method Overview.}
Figure 1 presents the overall architecture of MeshPriorDiT.
MeshPriorDiT decomposes future cloth motion into a structured mesh prior and a
generative residual. Given historical states and planned actions, the Mesh
Prior autoregressively predicts \(H\)-step displacements over the fixed
material topology. Residual DiT conditions on this prior and generates
displacement residuals through conditional flow matching. After
denormalization and scaling by \(\gamma\), the residuals are combined with the
prior and accumulated into raw future positions.

The topology-aware decoder then applies one row-normalized Jacobi update,
controlled by \(\lambda\), to the position residual relative to the online Mesh
Prior. The final trajectory is

\[
\widehat{\mathbf Q}_t
=
\mathbf F_H\odot
\left(
\widehat{\mathbf Q}_t^{\mathrm G}
+
\widehat{\mathbf E}_t^{\mathrm{topo}}
\right)
+
\mathbf M_H\odot\mathbf Q_t^{\mathrm{cmd}},
\qquad
\mathbf F=\mathbf 1-\mathbf M,
\]

where \(\widehat{\mathbf Q}_t^{\mathrm G}\) is the accumulated Mesh Prior
trajectory, \(\widehat{\mathbf E}_t^{\mathrm{topo}}\) is the decoded position
residual, and \(\mathbf Q_t^{\mathrm{cmd}}\) is the cumulative commanded
trajectory. This composition corrects free vertices while strictly anchoring
grasped vertices. The predicted positions are fed back as history during
long-horizon autoregressive rollout.

\subsection{Action-Conditioned Mesh Dynamics Prior}

\subsubsection{Graph Encoding and Constrained Prediction.}
The mesh prior propagates action-conditioned motion over the fixed material
topology \citep{longhini2025clothsplatting,pfaff2021meshgraphnets}. Given the
two most recent states, \(\mathbf q_{k-1}\) and \(\mathbf q_k\), each
undirected material connection is represented by two directed edges.

For vertex \(i\), the node feature is
\[
\begin{aligned}
\mathbf v_{k,i}^{\mathrm{act}}
&=
(1-M_i)(\mathbf q_{k,i}-\mathbf q_{k-1,i})
+M_i\mathbf a_{k+1},\\
\mathbf x_{k,i}
&=
[\mathbf v_{k,i}^{\mathrm{act}},
\operatorname{onehot}(c_i)],
\end{aligned}
\]
where \(c_i\) distinguishes free and grasped cloth vertices. For a directed
material edge \((i,j)\), we use
\[
\mathbf e_{k,ij}
=
\left[
\mathbf q_{k,i}-\mathbf q_{k,j},
\|\mathbf q_{k,i}-\mathbf q_{k,j}\|_2,
\mathbf u_i-\mathbf u_j,
\|\mathbf u_i-\mathbf u_j\|_2
\right],
\]
where \(\mathbf u_i=\mathbf U_{i,:}\) is the template material coordinate.
Thus, the edge features encode both the current three-dimensional deformation
and the fixed two-dimensional material relation.

Separate MLP encoders map the node and edge features into latent
representations, followed by \(L\) residual MeshGraphNets message-passing
blocks and a node decoder. The decoder predicts normalized discrete
accelerations, which are converted to physical scale and integrated as
\[
\begin{aligned}
\widehat{\overline{\boldsymbol\alpha}}_{k,i}
&=
\operatorname{Dec}(\mathbf h_i^L),\\
\widehat{\boldsymbol\alpha}_k
&=
\mathcal N_\alpha^{-1}
(\widehat{\overline{\boldsymbol\alpha}}_k),\\
\widetilde{\mathbf q}_{k+1}
&=
2\mathbf q_k-\mathbf q_{k-1}
+\widehat{\boldsymbol\alpha}_k.
\end{aligned}
\]

The planned action then overwrites the candidate prediction at grasped
vertices:
\[
\widehat{\mathbf q}_{k+1}
=
\mathbf F\odot\widetilde{\mathbf q}_{k+1}
+
\mathbf M\odot
(\mathbf q_k+\mathbf a_{k+1}^{\mathrm{vtx}}).
\]
This preserves the graph prediction at free vertices while enforcing the
prescribed grasp motion exactly.

\subsubsection{Prior Training and Multi-Step Rollout.}
The mesh prior is trained independently with normalized single-step
acceleration supervision. Let
\(\overline{\boldsymbol\alpha}_k^*\) denote the ground-truth target.
The free-vertex loss is
\[
\mathcal L_{\mathrm{prior}}(\theta_{\mathrm G})
=
\frac{
\sum_{i=1}^{N}
F_i
\left\|
\widehat{\overline{\boldsymbol\alpha}}_{k,i}
-
\overline{\boldsymbol\alpha}_{k,i}^{*}
\right\|_2^2
}{
\sum_{i=1}^{N}F_i
}.
\]

Although the full model receives \(K=3\) historical states, the second-order
mesh update uses only the two most recent states. Starting from
\(\widehat{\mathbf q}_{t-1}=\mathbf q_{t-1}\) and
\(\widehat{\mathbf q}_{t}=\mathbf q_t\), we recursively apply the constrained
update to obtain
\[
\widehat{\mathbf P}_t^{\mathrm G}
=
\left[
\widehat{\mathbf q}_{t+h}
-
\widehat{\mathbf q}_{t+h-1}
\right]_{h=1}^{H},
\qquad
\widehat{\mathbf q}_t=\mathbf q_t.
\]
This sequence serves as a topology-aware structured motion prior for
Residual DiT rather than the final prediction.

\subsection{Residual DiT}
\subsubsection{Prior-Conditioned Spatiotemporal Modeling.}
Residual DiT uses the mesh rollout as a reference and jointly models the residual between the ground-truth future increments and the \(H\)-step mesh prior. The residual in physical displacement space and its normalized target are defined as
\[
\begin{gathered}
\mathbf R_t^*=\Delta\mathbf Q_t^*-\widehat{\mathbf P}_t^{\mathrm G},\
\overline{\mathbf R}_t^*=\mathbf F_H\odot\operatorname{clip}\left(\mathbf R_t^*\oslash\mathbf S,-1,1\right).
\end{gathered}
\]

Here, \(\mathbf S\in\mathbb{R}^3\) is a fixed coordinate-wise scale computed from the training-set residual statistics and broadcast along the temporal and vertex dimensions. The operator \(\oslash\) denotes element-wise division, and \(\operatorname{clip}(\cdot,-1,1)\) denotes element-wise clipping. The mask \(\mathbf F_H\) is broadcast along the coordinate dimension, ensuring that the residual target is zero at grasped vertices.

We use a patch-token-based Diffusion Transformer (DiT) as the spatiotemporal generative backbone \citep{peebles2023dit}. The model takes \(K\) historical states and \(H\) steps of noisy residuals as input, with the grasp mask appended to every input frame. Planned actions are encoded as action tokens, while the mesh-prior tokens contain step-wise prior increments, cumulative prior displacements, and grasp indicators. The mesh prior therefore defines the residual target and also serves as an explicit condition, allowing DiT to jointly model residual motion across vertices and time over the full prediction horizon.

\subsubsection{Conditional Residual Flow Matching.}
We use conditional Flow Matching to learn the generative process of the normalized residual \citep{lipman2023flowmatching}. Gaussian noise is sampled at free vertices, and a linear probability path is constructed between the noise and the ground-truth residual:
\[
\begin{gathered}
\boldsymbol\epsilon=\mathbf F_H\odot\boldsymbol\epsilon_0,\quad\boldsymbol\epsilon_0\sim\mathcal N(\mathbf 0,\mathbf I),\\
\mathbf X_\tau=(1-\tau)\boldsymbol\epsilon+\tau\overline{\mathbf R}_t^*,\
\mathbf u_\tau^*=\overline{\mathbf R}_t^*-\boldsymbol\epsilon.
\end{gathered}
\]
Here, \(\tau\sim\mathcal U(0,1)\) is the flow time, and \(\mathbf u_\tau^*\) is the target velocity associated with the linear path.

Conditioned on the historical states, planned actions, mesh prior, and grasp mask, Residual DiT predicts a conditional velocity field. The effective velocity used during both training and inference is defined as

\[
\widehat{\mathbf v}_{t,\tau}=\mathbf F_H\odot\mathbf v_\phi\left(\mathbf X_\tau,\tau\mid\mathbf Q_t^-,\mathbf A_t,\widehat{\mathbf P}_t^{\mathrm G},\mathbf M\right).\]

Let \(N_{\mathrm F}=\sum_{i=1}^{N}F_i\) denote the number of free vertices. The velocity-matching loss is

\[\mathcal L_{\mathrm{flow}}(\phi)=\mathbb E_{\tau,\boldsymbol\epsilon}\left[\frac{\left\|\widehat{\mathbf v}_{t,\tau}-\mathbf u_\tau^*\right\|_2^2}{3HN_{\mathrm F}}\right].\]

During inference, the noise, intermediate flow states, and effective velocity are explicitly projected onto the free vertices. The residual generation process therefore does not modify the prescribed grasp motion.

Inference starts from the constrained Gaussian noise \(\mathbf X_0=\boldsymbol\epsilon\) and solves the conditional ordinary differential equation
\[
\begin{gathered}
\frac{\mathrm d\mathbf X_\tau}{\mathrm d\tau}=\widehat{\mathbf v}_{t,\tau},\
\widehat{\mathbf R}_t^{\mathrm{phy}}=\mathbf S\odot\mathbf X_1.
\end{gathered}
\]
We use the second-order Heun method for numerical integration. The number of integration steps and function evaluations is given in the implementation details. The endpoint \(\mathbf X_1\) is the generated normalized residual, and \(\widehat{\mathbf R}_t^{\mathrm{phy}}\) is its recovered form in physical displacement space.

\begin{figure*}[t]
\centering
\includegraphics[width=0.9\textwidth]{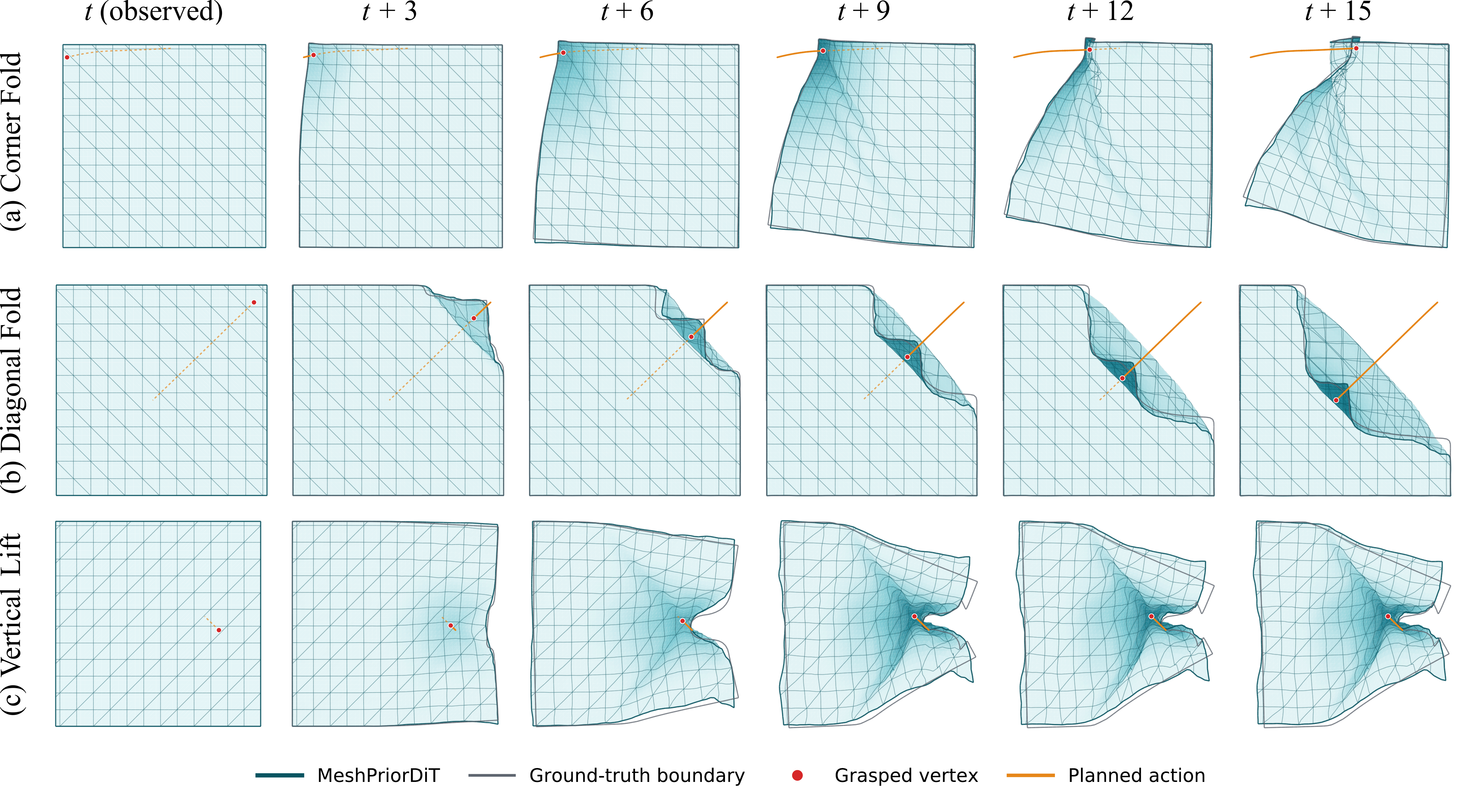} 
\caption{\textbf{Qualitative long-horizon rollouts of MeshPriorDiT.} Each row shows six snapshots of corner-to-side folding, diagonal folding, and vertical lifting. MeshPriorDiT autoregressively predicts cloth deformation from the initial state and planned actions. Blue and gray lines denote the prediction and ground-truth boundary, while red points and orange lines indicate grasped vertices and planned actions.}
\label{fig:Front}
\end{figure*}

\subsubsection{Topology-Aware Residual Decoding and Hierarchical Composition.}

Because the per-vertex Flow Matching objective does not explicitly enforce
spatial consistency, neighboring vertices may receive abruptly different
corrections. We therefore apply a fixed, parameter-free decoder to the
position residual relative to the online Mesh Prior.

Let \(\widehat{\mathbf r}_{t+h}^{\mathrm{phy}}\) be the denormalized
displacement residual. We first scale it by \(\gamma\) and obtain the raw
prediction:

\[
\begin{aligned}
\widehat{\Delta\mathbf q}_{t+h}^{\mathrm{raw}}
&=
\mathbf F\odot
\left(
\widehat{\mathbf p}_{t+h}^{\mathrm G}
+\gamma\widehat{\mathbf r}_{t+h}^{\mathrm{phy}}
\right)
+\mathbf M\odot\mathbf a_{t+h}^{\mathrm{vtx}},\\
\widehat{\mathbf q}_{t+h}^{\mathrm{raw}}
&=
\mathbf q_t+\sum_{j=1}^{h}
\widehat{\Delta\mathbf q}_{t+j}^{\mathrm{raw}},\\
\mathbf e_{t+h}^{\mathrm{raw}}
&=
\widehat{\mathbf q}_{t+h}^{\mathrm{raw}}
-\widehat{\mathbf q}_{t+h}^{\mathrm G}.
\end{aligned}
\]

Let \(\widetilde{\mathbf W}_{\mathcal G}
=\mathbf D^{-1}\mathbf W_{\mathcal G}\) be the row-normalized adjacency of the
fixed material mesh. A single Jacobi update gives

\[
\mathbf e_{t+h}^{\mathrm{topo}}
=
\mathbf F\odot
\left[
(1-\lambda)\mathbf e_{t+h}^{\mathrm{raw}}
+
\lambda\widetilde{\mathbf W}_{\mathcal G}
\mathbf e_{t+h}^{\mathrm{raw}}
\right].
\]

The final position is

\[
\widehat{\mathbf q}_{t+h}
=
\mathbf F\odot
\left(
\widehat{\mathbf q}_{t+h}^{\mathrm G}
+\mathbf e_{t+h}^{\mathrm{topo}}
\right)
+
\mathbf M\odot
\left(
\mathbf q_t+\sum_{j=1}^{h}
\mathbf a_{t+j}^{\mathrm{vtx}}
\right).
\]

Here, \(\gamma\) controls the residual magnitude and \(\lambda\) controls its
spatial coordination. Setting \(\gamma=0\) recovers the Mesh Prior, while
\(\gamma=1,\lambda=0\) uses the full unsmoothed residual. The grasped vertices
always follow the commanded trajectory exactly.

\subsection{Long-Horizon Autoregressive Rollout}

For a rollout of length \(T=WH\), MeshPriorDiT recursively applies its
\(H\)-step predictor over \(W\) non-overlapping windows. After each window,
the most recent \(K\) predicted states are used as the history for the next
action block:
\[
\widehat{\mathbf Q}_{t+H}^{-}
=
[\widehat{\mathbf q}_{t+H-K+1},\ldots,\widehat{\mathbf q}_{t+H}].
\]
Only the first window uses observed states; subsequent windows use model
predictions without future ground truth or teacher forcing. The mesh prior,
residual, and grasp constraints are recomputed in every window.

\begin{table*}[t]
\centering
\caption{\textbf{Results on the test set.} The best results are shown in bold, and the second-best results are underlined.}
\label{tab:baseline_comparison}
\begin{tabular}{lccccc}
\hline
Method
& 5-step Global
& 15-step Global
& 20-step Global
& 35-step Global
& 35-step Edge \\
\hline
DiT-DDPM
& 3.004
& 14.991
& 13.707
& 20.568
& 143.526 \\
GNN-Only
& 1.314
& 6.615
& 5.830
& 15.460
& 6.975 \\
MeshPriorDiT-Balanced
& \underline{0.755}
& \underline{3.743}
& \underline{3.360}
& \underline{10.372}
& \textbf{6.199} \\
MeshPriorDiT-MinMSE
& \textbf{0.662}
& \textbf{3.708}
& \textbf{3.294}
& \textbf{6.107}
& \underline{6.898} \\
\hline
\end{tabular}
\end{table*}

\section{Experiments}
We organize our experiments around four research questions:\\
\textbf{Q1:} Does MeshPriorDiT improve action-conditioned cloth dynamics prediction over GNN-based structured and DDPM-based generative baselines?\\
\textbf{Q2:} Does MeshPriorDiT retain its advantage under long-horizon autoregressive rollouts?\\
\textbf{Q3:} Does the mesh prior provide a topology-aware physical reference, and how does Residual DiT model the remaining dynamics?\\
\textbf{Q4:} How do residual gain and topology-aware decoding affect the trade-off between trajectory accuracy and local mesh quality?

\subsection{Experimental Setup}

\paragraph{Dataset and Splits.}
We construct a 500,000-window cloth dynamics dataset in SoftGym. Complete simulation trajectories are first divided into 24,822/3,106/3,106 disjoint train/validation/test sets, yielding 400,000/50,000/50,000 windows with three observed and five target frames. Thus, no trajectory or overlapping window crosses splits. Diagonal Fold, Corner-to-Side Fold, and Vertical Lift are approximately balanced within each split.

\paragraph{Baselines and Operating Points.}
We compare MeshPriorDiT with GNN-Only and a reproduced UniClothDiff baseline, denoted DiT-DDPM. GNN-Only uses the same mesh prior and rollout protocol but removes residual correction (\(\gamma=0\)), while the \(\gamma\)–\(\lambda\) sweep evaluates residual scaling and topology-aware smoothing. Using one checkpoint, we select two operating points exclusively on validation data: Balanced (\(\gamma=0.5,\lambda=0.8\)) trades off Global and Edge-strain MSE, whereas MinMSE (\(\gamma=1.0,\lambda=0.8\)) minimizes coordinate error.

\paragraph{Training and Inference.}
The Mesh Prior contains 15 message-passing blocks with a hidden dimension of 128 and is trained independently. The Residual DiT is trained on 16 NVIDIA A100 GPUs with a global batch size of 2,048, AdamW, and BF16 mixed precision. The Mesh Prior remains frozen during this stage, and only the original Flow Matching objective is optimized. At inference time, we use a 50-step Heun solver. Since each solver step requires one prediction and one correction evaluation, each five-frame window requires 100 neural function evaluations (NFE100).

\paragraph{Metrics.}
All metrics are computed after restoring predictions to physical coordinate space. We report Global MSE and Edge-strain MSE, where lower values indicate better performance. All relative improvements are computed from the original unrounded evaluation records. Qualitative 15-step rollouts are shown in Figure 2.

\subsection{Q1: Comparison with Baselines}

We first examine whether MeshPriorDiT outperforms the GNN-Only and DiT-DDPM baselines. Table 1 reports Global MSE across different rollout horizons and Edge-strain MSE on the test set. Coordinate errors are reported in units of \(10^{-5}\), while Edge-strain MSE is reported in units of \(10^{-3}\).

For the 5-step, 15-step, 20-step, and 35-step rollouts, MeshPriorDiT-MinMSE reduces Global MSE relative to GNN-Only by \(49.59\%\), \(43.94\%\), \(43.49\%\), and \(60.50\%\), respectively. Relative to DiT-DDPM, the corresponding reductions are \(77.95\%\), \(75.26\%\), \(75.97\%\), and \(70.31\%\). MeshPriorDiT-Balanced also consistently outperforms GNN-Only across all four rollout horizons, reducing Global MSE by \(42.57\%\), \(43.42\%\), \(42.36\%\), and \(32.91\%\), respectively. Thus, both operating points consistently improve overall coordinate accuracy across different prediction horizons.

At the longest 35-step rollout horizon, Balanced achieves the lowest Edge-strain MSE, reducing it by approximately \(11.1\%\) relative to GNN-Only. In contrast, MinMSE achieves the lowest coordinate error, but its Edge-strain MSE is slightly higher than that of the Balanced configuration. This result indicates that the magnitude of residual correction introduces a trade-off between trajectory accuracy and local mesh quality, which is further analyzed in Q4.

\subsubsection{Answer to Q1.}
MeshPriorDiT outperforms both GNN-Only and DiT-DDPM across all evaluated rollout horizons. The results support the complementarity between the mesh prior and Residual DiT: GNN-Only provides a low-error structured motion reference grounded in the material topology, while Residual DiT further corrects motion discrepancies that are not accurately captured by the prior.

\begin{figure}[!htbp]
\centering
\includegraphics[width=0.9\columnwidth]{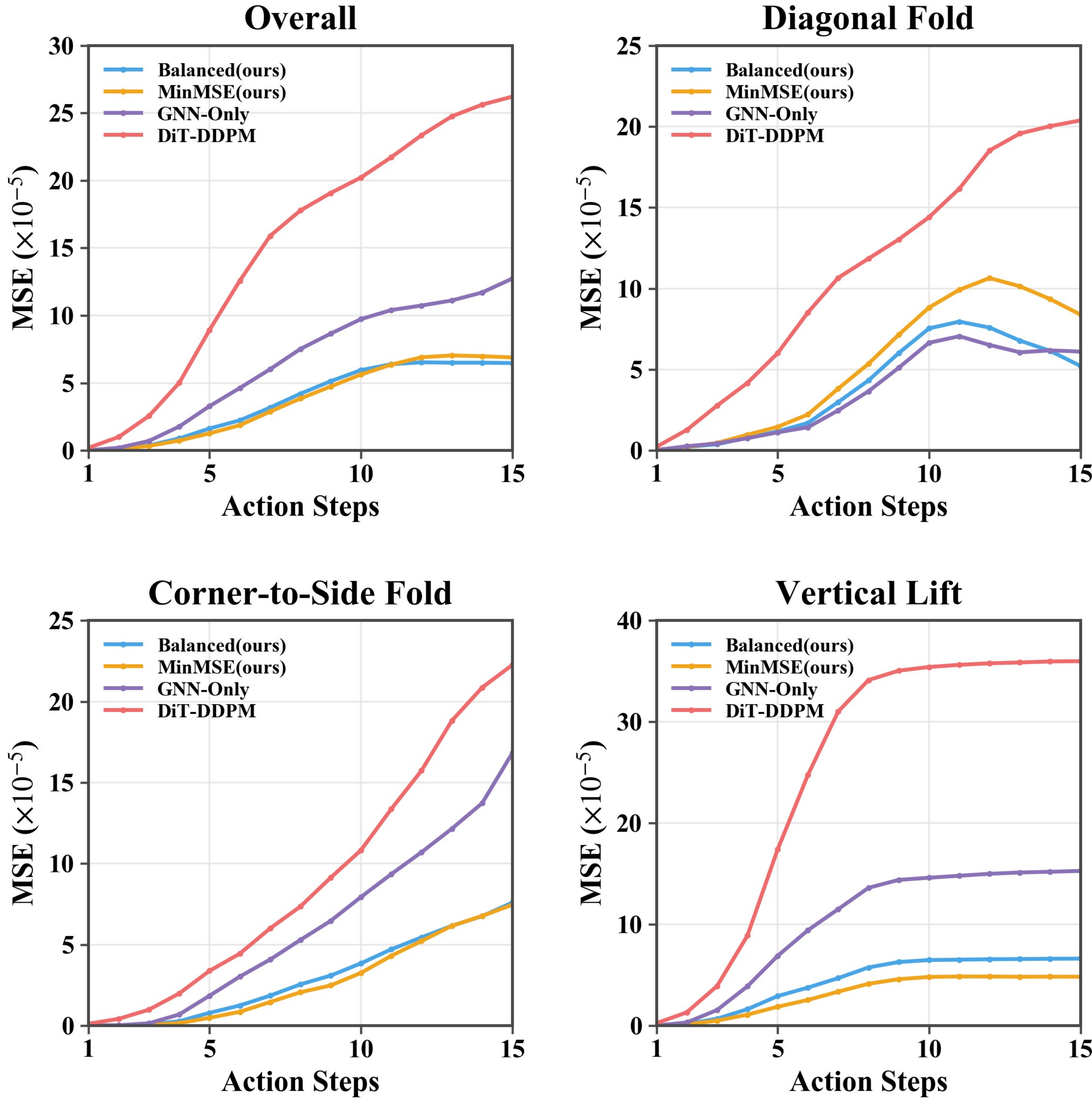} 
\caption{Per-step Global MSE over 15-step autoregressive TEST rollouts, averaged over 60 trajectories overall and 20 per action. Lower is better.}
\label{fig:loss}
\end{figure}

\subsection{Q2: Long-Horizon Autoregressive Rollout}
We further investigate whether the performance gains of MeshPriorDiT persist under long-horizon autoregressive rollouts, where the predicted states from each five-step window are fed back as the historical input to the next window.

As shown in Figure 3, the errors of GNN-Only and DiT-DDPM progressively accumulate during autoregressive rollout, whereas Balanced (ours) and MinMSE (ours) maintain substantially lower errors. At the 15-step horizon, Balanced and MinMSE reduce the average Global MSE by \(43.42\%\) and \(43.94\%\) relative to GNN-Only, and by \(75.03\%\) and \(75.26\%\) relative to DiT-DDPM, respectively.

The task-wise results further show that, for Corner-to-Side Fold and Vertical Lift, both MeshPriorDiT configurations exhibit substantially slower error growth than the baselines. Relative to GNN-Only, Balanced reduces the H15 Global MSE by \(51.89\%\) and \(56.98\%\) on the two tasks, respectively, while MinMSE achieves reductions of \(55.85\%\) and \(68.86\%\). On Diagonal Fold, where GNN-Only already provides a strong prediction, residual correction offers limited benefit and can slightly degrade the aggregate error. Nevertheless, it still maintains very good results.

\subsubsection{Answer to Q2.}
The advantages of MeshPriorDiT over GNN-Only and DiT-DDPM persist throughout autoregressive rollout, indicating that Residual DiT remains effective in correcting the accumulated coordinate drift of the Mesh Prior after multiple rounds of autoregressive feedback.

\begin{figure}[!htbp]
\centering
\includegraphics[width=\columnwidth]{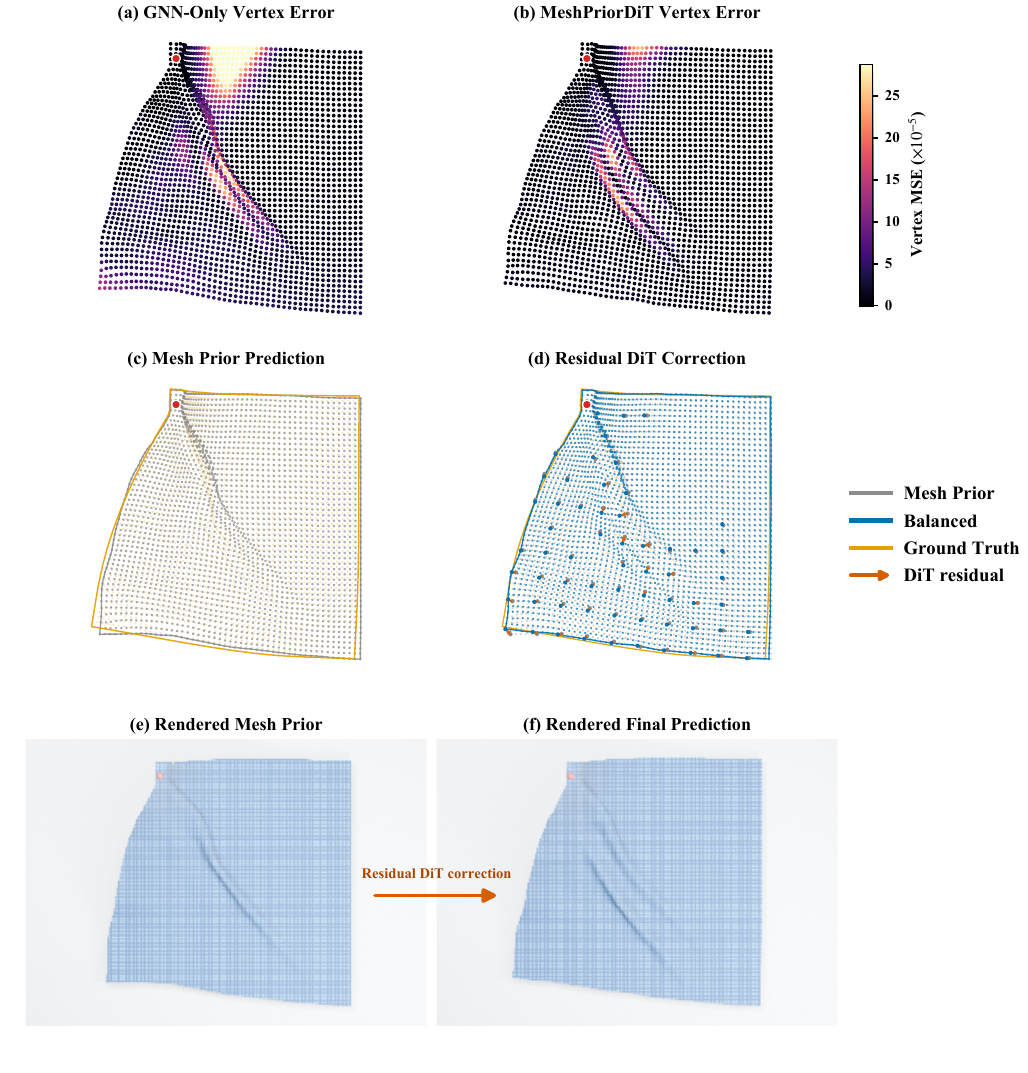} 
\caption{\textbf{Roles of the mesh prior and Residual DiT.}
(a--b) Vertex errors before and after residual correction.
(c--d) The residual corrects local deviations of the mesh prior.
(e--f) Rendered prior and final predictions.
}
\label{fig:mesh-residual}
\end{figure}

\subsection{Q3: Roles of the Mesh Prior and Residual DiT}
To examine the roles of the two branches, we compare DiT-DDPM, GNN-Only, and the complete model. Table 1 shows that GNN-Only outperforms DiT-DDPM, indicating that mesh topology, local connectivity, and grasp constraints provide effective physical inductive biases. The complete model further improves upon GNN-Only, demonstrating that the residual branch compensates for errors in the mesh prior.

Figure 4 further illustrates this division of responsibilities. Figure 4(a) shows that the larger errors of GNN-Only are concentrated near the grasped vertex and along the folding region extending downward from it. After introducing Residual DiT, the high-error region in Figure 4(b) becomes substantially smaller and more localized around complex deformations. Figure 4(c) shows that the mesh prior captures the overall boundary and primary folding structure but retains local deviations from the ground truth. In Figure 4(d), the generated residuals primarily act on these regions, bringing the composed prediction closer to the ground-truth mesh.

\subsubsection{Answer to Q3.}
The mesh prior captures the primary motion and local physical propagation under topological constraints, while Residual DiT provides targeted corrections in folding and highly deformable regions. Together, they form a complementary hierarchical prediction mechanism.

\begin{figure}[!htbp]
\centering
\includegraphics[width=\columnwidth]{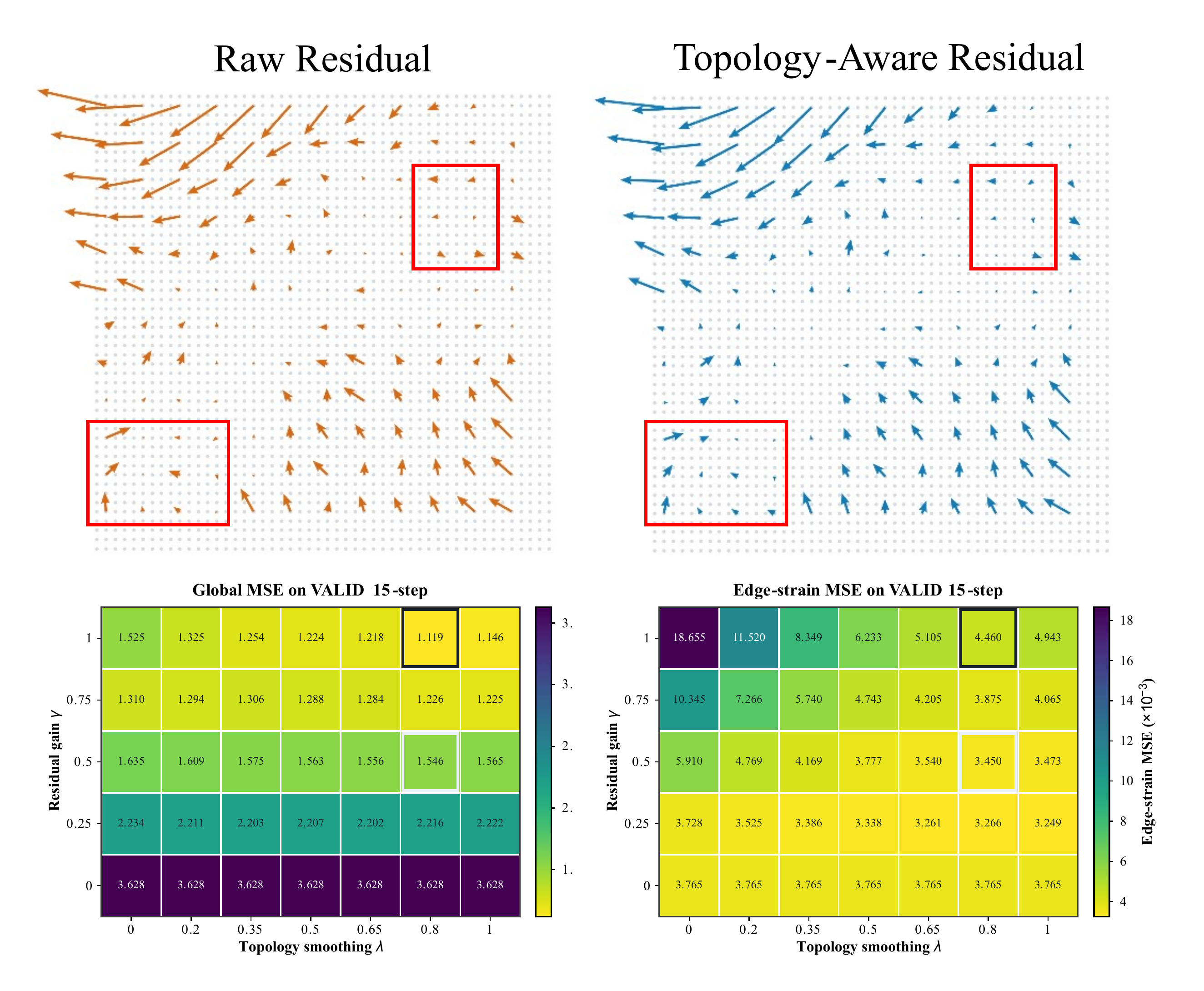} 
\caption{\textbf{Topology-aware residual decoding and parameter sensitivity.} The top row visualizes the raw generated residual and its topology-aware counterpart after material-neighborhood smoothing. The bottom row reports 15-step Global MSE and Edge-strain MSE under different residual gains \(\gamma\) and smoothing strengths \(\lambda\). The outlined cells indicate the selected configuration.}
\label{fig:s-lamda-residual}
\end{figure}

\subsection{Q4: Balancing Trajectory Accuracy and Mesh Quality}

We use the residual gain \(\gamma\) to control the magnitude of the DiT correction and \(\lambda\) to apply one step of Jacobi residual smoothing over the fixed template adjacency, without altering the original mesh topology.

The top row of Figure 5 shows Vertical Lift residuals at step 10, projected onto the \(x\)-\(z\) plane and enlarged \(4\times\); red boxes mark the same neighborhood. Smoothing improves local consistency, reducing Neighbor Variation, Edge-strain MSE, and Coordinate MSE by \(80.4\%\), \(78.1\%\), and \(12.3\%\).

The second row of Figure 5 shows that increasing \(\gamma\) reduces Global MSE but amplifies local edge strain, whereas moderately increasing \(\lambda\) improves both metrics. At \(\gamma=1\), setting \(\lambda=0.8\) reduces Global MSE from \(1.525\) to \(1.119\) and Edge-strain MSE from \(18.655\) to \(4.460\), while stronger smoothing leads to slight degradation.

\subsubsection{Answer to Q4.}
The residual gain \(\gamma\) controls the correction strength, while \(\lambda\) suppresses local high-frequency residuals. The Balanced configuration prioritizes a trade-off between trajectory accuracy and mesh quality, whereas MinMSE prioritizes the lowest coordinate error.

\section{Conclusion}
We presented MeshPriorDiT, which decomposes action-conditioned cloth
dynamics into a topology-aware mesh prior and a generative residual. The
mesh GNN captures structured local motion, while Residual DiT corrects
the remaining trajectory errors, particularly in highly deformable regions.
Across all three tasks at 15-step, MeshPriorDiT reduces aggregate Global MSE by \(43.42\%\) and \(75.03\%\) relative to GNN-Only and DiT-DDPM, respectively. At 35-step, it achieves reductions of \(32.91\%\) and \(49.57\%\), respectively. These results demonstrate that MeshPriorDiT maintains effective residual correction over long autoregressive rollouts while balancing trajectory accuracy and local structural consistency. Future work will investigate varying material properties and real-world cloth manipulation.

\bibliography{references}
\end{document}